\documentclass{article} 
\usepackage{iclr2026_conference,times}

\usepackage{amsmath,amsfonts,bm}

\def\eqref#1{equation~\ref{#1}}

\def\1{\bm{1}}

\DeclareMathAlphabet{\mathsfit}{\encodingdefault}{\sfdefault}{m}{sl}
\SetMathAlphabet{\mathsfit}{bold}{\encodingdefault}{\sfdefault}{bx}{n}

\usepackage{hyperref}
\usepackage{url}
\usepackage{graphicx}
\usepackage{listings}
\usepackage{placeins}
\usepackage{booktabs}
\usepackage{float}

\title{Automated Concept Discovery for LLM-as-a-Judge Preference Analysis}

\author{James Wedgwood, Chhavi Yadav, \& Virginia Smith \\
Department of Computer Science\\
Carnegie-Mellon University\\
Pittsburgh, PA 15213, USA \\
\texttt{\{jwedgwoo,cyadav,smithv\}@cs.cmu.edu} \\
}

\iclrfinalcopy 
\begin{document}

\maketitle

\begin{abstract}
Large Language Models (LLMs) are increasingly used as scalable evaluators of model outputs, but their preference judgments exhibit systematic biases and can diverge from human evaluations. Prior work on LLM-as-a-judge has largely focused on a small, predefined set of hypothesized biases, leaving open the problem of automatically discovering unknown drivers of LLM preferences. We address this gap by studying several embedding-level concept extraction methods for analyzing LLM judge behavior. We compare these methods in terms of interpretability and predictiveness, finding that sparse autoencoder–based approaches recover substantially more interpretable preference features than alternatives while remaining competitive in predicting LLM decisions. Using over 27k paired responses from multiple human preference datasets and judgments from three LLMs, we analyze LLM judgments and compare them to those of human annotators. Our method both validates existing results, such as the tendency for LLMs to prefer refusal of sensitive requests at higher rates than humans, and uncovers new trends across both general and domain-specific datasets, including biases toward responses that emphasize concreteness and empathy in approaching new situations, toward detail and formality in academic advice, and against legal guidance that promotes active steps like calling police and filing lawsuits. Our results show that automated concept discovery enables systematic analysis of LLM judge preferences without predefined bias taxonomies.
\end{abstract}

\section{Introduction}

High-quality evaluation of language model responses is crucial for improving performance, but human feedback can be costly and difficult to obtain. Recent work has shown that modern LLMs can provide scalable preference judgments, motivating a large literature on using LLMs as evaluators of model outputs. LLM-as-a-judge was formalized and stress-tested on response-quality tasks by \citet{zheng2023judgingllmasajudgemtbenchchatbot}, including evidence of systematic biases such as position and self-enhancement effects. Most follow-on studies investigate a fixed, known set of biases or factors, treating discovery as manual hypothesis testing rather than automated concept exploration. This leaves a gap in tools for uncovering unknown preference drivers, including those that may only surface in narrow or specialized domains.

In this paper, we apply a suite of concept discovery techniques to analyze the preference patterns of LLM judges, including both supervised and unsupervised methods. We focus on techniques that take prompt and response embeddings as input, producing features that encode axes of difference between chosen and rejected responses, such as responding to versus refusing a request or providing specific versus general answers. Using a composite dataset drawn from three high-quality human preference corpora, we obtain preference judgments from three recent strong models from different providers: OpenAI's \texttt{gpt-5.1}, Anthropic's \texttt{claude-sonnet-4.5}, and Google's \texttt{gemini-3-flash-preview}. Features are generated via both supervised and unsupervised methods, and an automated interpretability pipeline is applied to construct descriptions for these features, which are validated for fidelity against held-out samples. With these descriptions in hand, we analyze the impact of these difference axes on LLM preference, focusing in particular on cases where LLM judgments differ significantly from those of humans.

The concept extraction methods used in our paper include classical techniques based on principal component analysis (PCA), as well as more modern techniques leveraging sparse autoencoders (SAEs). Automated concept discovery via SAEs has recently been framed for human feedback analysis by \citet{movva2025whatshumanfeedbacklearning}, where small SAEs are trained on the difference between response embeddings to uncover interpretable preference features. Unlike these human feedback settings, where annotators for distinct datasets may exhibit very different preference patterns, LLM-as-a-judge provides a shared evaluator across many datasets, enabling a single SAE to be trained jointly on heterogeneous corpora. This creates a more statistically efficient and transferable discovery setting than bespoke per-dataset encoders.

The main results of this paper are twofold. First, we compare the strengths and weaknesses of different concept extraction methods using proxy metrics for interpretability (how many features with high-fidelity interpretations does each method yield) and predictiveness (how well are the generated features able to predict LLM judgments). We find that supervised methods are much more predictive than unsupervised ones, yielding up to a 138\% increase in predictiveness versus the best unsupervised methods when compared to random guessing; however, this comes at a steep cost to interpretability. We also find that SAE-based methods yield far more interpretable features than PCA with little to no decrease in predictiveness.

Next, we use the generated feature descriptions to comprehensively analyze LLM judgment factors. We find that LLMs prefer refusal of sensitive requests at rates higher than humans, validating previous results \citep{pasch2025aivshumanjudgment}; in particular, \texttt{claude-sonnet-4.5} errs strongly on the side of responses that encode refusal or AI limitations. We also explore previously-unknown preference drivers, finding that LLMs are more likely to prefer responses that emphasize measurability, concreteness, empathy, and emotions, while humans tend to value flexibility, uncertainty, and personal growth. Our methods also allow for systematic bias mining even in niche cases where latent, field-specific, or otherwise unexpected preference drivers may exist but lack predefined taxonomies for manual study. To this end, we analyze datasets related to academic and legal advice, uncovering biases toward detailed, formal responses to academic questions, and against legal guidance that directs users to external resources or encourages them to take matters into their own hands through steps like lawsuits, surveillance technology, and involving police.

\section{Related Work}

\paragraph{LLM-as-a-judge preference analysis.} A large volume of work has been produced with the aim of better understanding the basis of LLM judge preferences. Section 4 of \citet{gu2025surveyllmasajudge} provides a helpful survey. Well-known patterns such as position bias, self-enhancement bias, and verbosity bias were first studied by \citet{zheng2023judgingllmasajudgemtbenchchatbot} and elaborated in later work, e.g. \citet{ye2024justiceprejudicequantifyingbiases}, \citet{shi2025judgingjudgessystematicstudy}, \citet{huang2025empiricalstudyllmasajudgellm}. The present paper builds on this work by using concept extraction to discover previously unknown sources of bias, as opposed to identifying suspected biases and then testing for those.

\paragraph{LLM versus human preferences.} Along with the literature dealing specifically with LLM judge preferences, there is substantial work comparing LLMs to human annotators. \citet{thakur2025judgingjudgesevaluatingalignment} proposes metrics for human–LLM alignment beyond the basic percent agreement. \citet{movva2024annotationalignmentcomparingllm} and \citet{pasch2025aivshumanjudgment} compare judgments in the context of safety annotations and content moderation, respectively. \citet{li2024dissectinghumanllmpreferences}, \citet{oh2025uncoveringfactorlevelpreferences}, and \citet{chen2024humansllmsjudgestudy} all identify specific preference factors that differ between LLM and human judges; again, in contrast to this work, all of these papers do so by hand-selecting and then investigating known factors.

\paragraph{Concept extraction and SAEs.} Automated concept extraction is a wide-ranging and well-established field of machine learning; \citet{fel2023holisticapproachunifyingautomatic} provides a helpful conceptual framework for comparing different techniques. This paper uses sparse autoencoders (SAEs) for concept extraction from embeddings, a technique first developed in \citet{oneill2024disentanglingdenseembeddingssparse} and applied in the setting of human preference data by \citet{movva2025whatshumanfeedbacklearning}, which provides the basis for much of our method. Several recent papers advocate for the usage of SAEs for topic modeling and concept discovery beyond their more well-known role as a mechanistic interpretability tool, including \citet{peng2025usesparseautoencodersdiscover} and \citet{girrbach2025sparseautoencoderstopicmodels}.

\section{Methodology}

Our methods seek to address the following research questions:
\begin{itemize}
    \item \textbf{RQ1.} Which embedding-level concept extraction techniques are most successful at yielding interpretable preference axes and at predicting LLM judgments?
    \item \textbf{RQ2.} Can we use concept extraction to validate known biases of LLM judges?
    \item \textbf{RQ3.} Can we use concept extraction to discover previously unknown preference drivers for LLM judges, especially in cases where they differ from humans?
\end{itemize}

\subsection{Data Preparation}

To create a setting where LLM judge preferences can be analyzed consistently across a variety of inputs, we combine three well-known human preference datasets and standardize them into a binarized form, randomly assigning one response from each pair to be $r_A$ and the other to be $r_B$. The datasets are Community Alignment \citep{zhang2025cultivating}, LMArena 100k \citep{chiang2024chatbot}, and PRISM \citep{kirk2024PRISMdataset}. We follow previous work \citep{huang2025valueswilddiscoveringanalyzing} in removing prompts that require an objectively correct answer; in such cases, the primary preference criterion will be whether or not the answer is correct, which is unlikely to be analyzable by concept extraction techniques. For domain-specific analysis, we use the \texttt{askacademia} and \texttt{legaladvice} domains from SHP-2 \citep{pmlr-v162-ethayarajh22a}. Our composite dataset contains 27,734 entries and our domain-specific datasets have 10,418 total entries. For each response pair, we generate binary preference judgments using \texttt{gpt-5.1}, \texttt{claude-sonnet-4.5}, and \texttt{gemini-3-flash-preview}. More details on data preparation can be found in Appendix~\ref{app:method-details}, and LLM agreement and position bias statistics can be found in Appendix~\ref{app:model-agreement}.

\subsection{Concept Extraction}

\begin{figure}[t]
\begin{center}
\includegraphics[width=\linewidth]{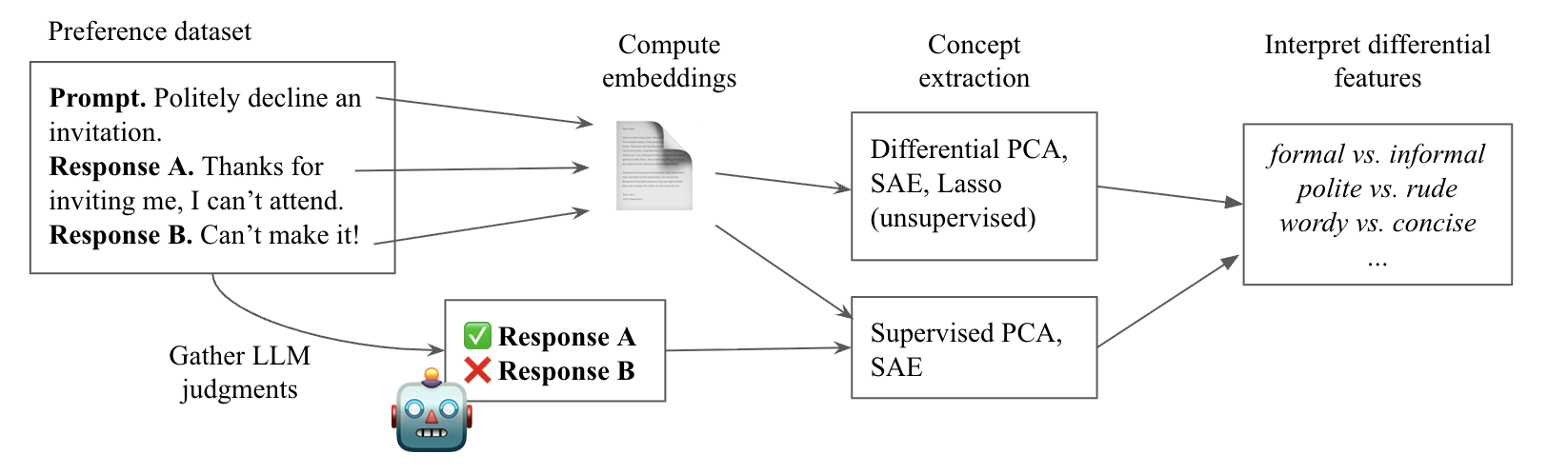}
\end{center}
\caption{High-level overview of methodology. Embeddings and LLM judgments are generated from a paired preference dataset, differential features are extracted, and interpretations of these features are generated for further study.}
\label{fig:flowchart}
\end{figure}

The end-to-end process of generating interpretable preference features via embedding-level concept extraction is diagrammed in Figure~\ref{fig:flowchart}. We first embed the prompt and both responses for each entry in the dataset using OpenAI's \texttt{text-embedding-3-small}. Several techniques are employed to produce concept features, with each feature representing an axis of difference along which a given response pair lies. For ease of comparison, we standardize our methods so that each one yields 32 total features. The methods we tested are as follows:
\begin{itemize}
    \item Differential PCA: A PCA model is fitted directly to the differences between response embeddings and the first 32 features are interpreted.
    \item Differential SAE: An SAE with 32 latents is trained on the difference between response embeddings and the features are interpreted. This is the method applied by \citet{movva2025whatshumanfeedbacklearning} to human preference data; for consistency, we use the same SAE design ($m=32,k=4$, Matryoshka BatchTopK SAE with prefixes $[8, 32]$). See \citet{bussmann2025learningmultilevelfeaturesmatryoshka} and \citet{bussmann2024batchtopksparseautoencoders} for Matryoshka and BatchTopK SAEs, respectively.
    \item Differential SAE + Lasso: A larger SAE (128 latents) is trained on the difference between response embeddings, then Lasso regression is used to select the 32 latents that are most predictive of the target LLM judgment.
    \item Supervised PCA: A neural network is trained on the prompt and response embeddings, with LLM preference as the target variable. A PCA model is fitted on the penultimate layer and the first 32 features are interpreted. This method and the next follow previous work on concept bottleneck models \citep{koh2020conceptbottleneckmodels} and implementation details may be found in Appendix~\ref{app:method-details}.
    \item Supervised SAE: A 32-latent SAE is trained on the penultimate layer of the same neural network from the previous method and its features are interpreted.
\end{itemize}

\subsection{Feature Interpretation}

We follow prior work \citep{bills2023language} to generate interpretable descriptions for these features. For a given feature $f$, let $a_i \in \mathbb{R}$ denote its signed activation on dataset entry $i$, where each entry consists of a paired comparison between $r_A$ and $r_B$. We select the five entries with the largest absolute activations $|a_i|$ and prompt \texttt{gpt-5.1} to propose a natural-language explanation of the latent difference axis represented by $f$, based on these examples. All descriptions are written from the perspective of $r_A$; for example, a feature described as \textit{provides a context-specific, substantive reply instead of a brief generic refusal} is expected to have $a_i > 0$ when $r_A$ is a substantive reply and $r_B$ is a refusal, and $a_i < 0$ in the reverse case.

To validate these descriptions, we randomly draw a held-out set of 100 additional entries with large $|a_i|$ for the same feature. For each entry, we prompt \texttt{gpt-5-mini} to indicate whether Response A, Response B, or neither more strongly exhibits the described feature. If the interpretation is faithful, the model should preferentially select Response A when $a_i > 0$ and Response B when $a_i < 0$. We quantify this alignment by testing the association between the model’s categorical choices and the signed activations $\{a_i\}$ using a permutation test. A feature is deemed interpretable if this test yields a Bonferroni-corrected $p$-value below $0.05$. This procedure closely follows that of \citet{movva2025whatshumanfeedbacklearning}; implementation details are provided in Appendix~\ref{app:method-details}, and full prompts are provided in Appendix~\ref{app:prompts}.

\section{Results}

\subsection{Method Comparison}

\begin{table}[t]

\caption{Interpretability and predictiveness of feature learning methods on combined preference dataset. Models: GPT = \texttt{gpt-5.1}, Claude = \texttt{claude-sonnet-4.5}, Gemini = \texttt{gemini-3-flash-preview}. Metric descriptions appear in the text body. For the last three methods, Interpretability varies depending on the model's target variable; a range is shown.}
\label{tab:interpretability-predictiveness}

\begin{center}
\begin{tabular}{lcccc}
\toprule
\multicolumn{1}{c}{\bf Method} &
\multicolumn{1}{c}{\bf Interp.} &
\multicolumn{1}{c}{\bf Pred. (GPT)} &
\multicolumn{1}{c}{\bf Pred. (Claude)} &
\multicolumn{1}{c}{\bf Pred. (Gemini)}
\\ \midrule 
Differential PCA        & 4     & 0.63 & 0.66 & 0.67 \\
Differential SAE        & \bf 18    & 0.61 & 0.65 & 0.66 \\
Diff.\ SAE + Lasso      & 7--17 & 0.63 & 0.66 & 0.67 \\
Supervised PCA          & 0--4  & \bf 0.81 & \bf 0.84 & \bf 0.84 \\
Supervised SAE          & 4--6  & \bf 0.81 & \bf 0.84 & 0.83 \\
\bottomrule
\end{tabular}
\end{center}

\end{table}

The desired qualities of a feature extraction technique in this setting are (1) interpretability, i.e. how many of the differential features admit high-fidelity, human-readable descriptions; (2) predictiveness, i.e. how well the identified features predict the preferences of an LLM judge. A summary of the techniques used with associated metrics appears in Table~\ref{tab:interpretability-predictiveness}. The Interpretability metric is simply the number of features for which an interpretable description was generated that met the significance criterion described above, out of a maximum of 32 for all techniques. The Predictiveness metric, meanwhile, represents the ROC-AUC of a logistic regression model fitted to the generated features with LLM preference as the target variable; see Appendix~\ref{app:method-details} for more details.

Due to the sparsity constraint, SAE-based methods have much higher Interpretability scores than other methods, with the Differential SAE producing over four times as many interpretable features as its PCA counterpart. Notably, using PCA instead of SAEs drastically reduces Interpretability with only slight gains in Predictiveness; the same is true of the Lasso-based method. Supervised methods, on the other hand, are far more predictive than unsupervised ones. Using random guessing (Predictiveness = 0.5) as a baseline, the Supervised PCA and SAE methods yield a 138\% improvement over the best unsupervised methods in Predictiveness of \texttt{gpt-5.1} preference, indicating that the relation between embedding differences and LLM judgments exhibits significant nonlinearity. We next turn to the features generated by the most interpretable method, the Differential SAE, to study LLM judge preferences and divergences from humans.

\subsection{Differential SAE Feature Analysis}

\begin{figure}[t]

\begin{center}
\includegraphics[width=\linewidth]{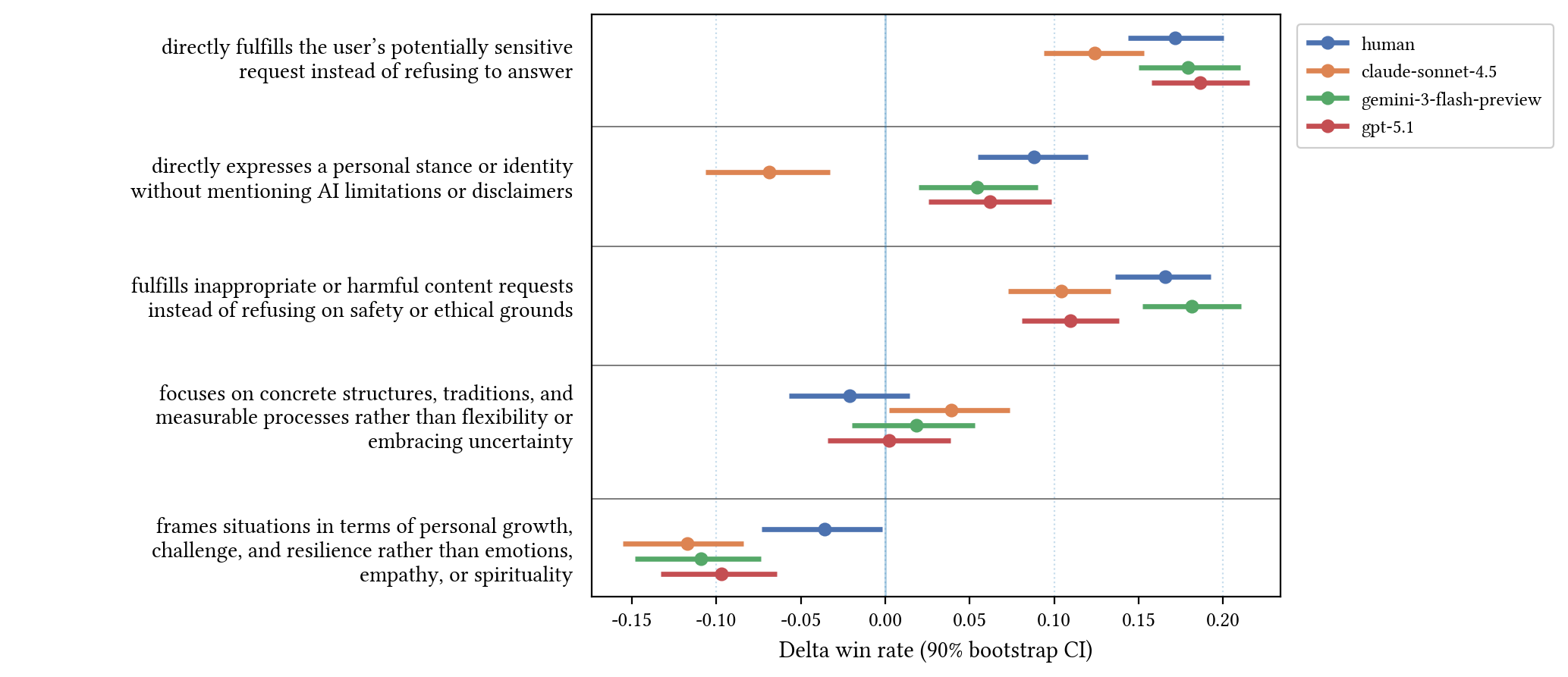}
\end{center}

\caption{A selection of Differential SAE features for the combined preference dataset, with interpretations and $\Delta$win-rate for human and LLM annotators.}
\label{fig:combined} 

\end{figure}

\paragraph{General analysis.} To assess the impact of response difference axes on preferences, we use the length-controlled $\Delta$win-rate metric \citep{movva2025whatshumanfeedbacklearning}, representing the predicted difference in win rate for positive versus negative values of the feature $f$, using logistic regression while holding other features constant. Intuitively, a feature with positive $\Delta$win-rate is generally preferred, while one with negative $\Delta$win-rate is generally dispreferred. Out of a total of 18 Differential SAE features with high-fidelity interpretations, we focus on a selection for which $\Delta$win-rate varies significantly between human versus LLM judges, as shown in Figure~\ref{fig:combined}. In Appendix~\ref{app:feature-analysis}, we provide a similar visualization with all features, and metric details can be found in Appendix~\ref{app:delta-winrate}.

Previous work \citep{pasch2025aivshumanjudgment} has shown that LLMs are more likely than humans to prefer refusal of sensitive prompts. This pattern is strongest for \texttt{claude-sonnet-4.5}, which has $\Delta$win-rate about four percentage points lower than humans on the feature \textit{directly fulfills the user's potentially sensitive request instead of refusing to answer}, while other LLMs more closely match humans. On the feature \textit{directly expresses a personal stance or identity without mentioning AI limitations or disclaimers}, meanwhile, \texttt{claude-sonnet-4.5} is the only judge to have a negative $\Delta$win-rate. On the feature \textit{fulfills inappropriate or harmful content requests instead of refusing on safety or ethical grounds}, however, both \texttt{claude-sonnet-4.5} and \texttt{gpt-5.1} have $\Delta$win-rate about six percentage points lower than humans.

We also recover several additional features for which significant discrepancies exist between LLM and human judges. On the feature \textit{focuses on concrete structures, traditions, and measurable processes rather than flexibility or embracing uncertainty}, $\Delta$win-rate is positive for all three models but negative for humans. For \textit{frames situations in terms of personal growth, challenge, and resilience rather than emotions, empathy, or spirituality}, meanwhile, $\Delta$win-rate is negative for both humans and LLMs, but lies at least seven percentage points higher for humans, suggesting different patterns of preference in approaching new situations.

Our method may also be used to shed further light on known tendencies, including self-enhancement bias, the pattern whereby LLMs prefer responses generated by similar models, studied by \citet{zheng2023judgingllmasajudgemtbenchchatbot} and \citet{ye2024justiceprejudicequantifyingbiases}. We find evidence of self-enhancement bias in \texttt{gpt-5.1}, which prefers responses from OpenAI models at a rate 12 percentage points higher than humans. In about 25\% of cases where \texttt{gpt-5.1} prefers the OpenAI response and humans do not, we find the feature \textit{fulfills inappropriate or harmful content requests instead of refusing on safety or ethical grounds} to be active for the OpenAI response, likely explaining some of the discrepancy.

\begin{figure}[t]

\begin{center}
\includegraphics[width=\linewidth]{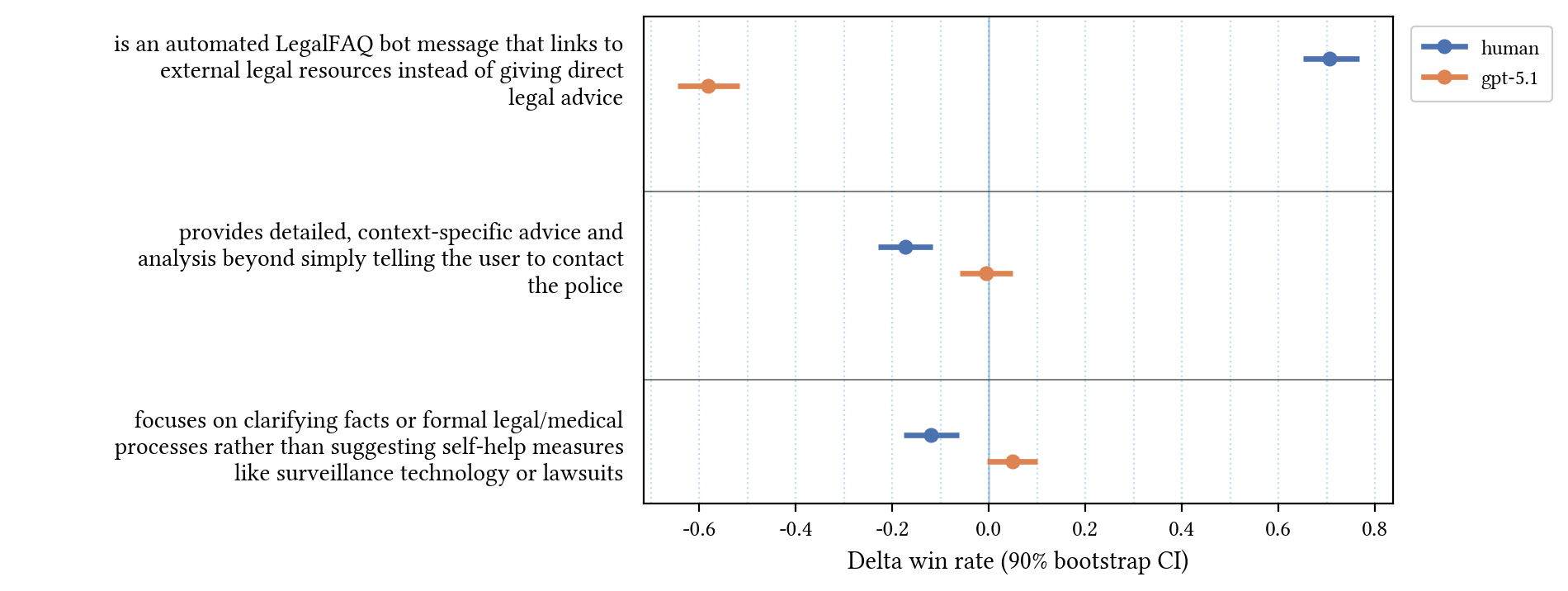}
\end{center}

\caption{Selected features for the \texttt{legaladvice} dataset, with interpretations and $\Delta$win-rate.}
\label{fig:legaladvice}

\end{figure}
\paragraph{Domain-specific dataset analysis.} To demonstrate the applicability of our method to domain-specific datasets, we perform a similar analysis on the \texttt{askacademia} and \texttt{legaladvice} domains from SHP-2 \citep{pmlr-v162-ethayarajh22a}, which draws preference data from Reddit posts. This time, judgments are generated by \texttt{gpt-5.1} only. Unlike on the combined preference dataset, where humans and LLMs agree roughly 70\% of the time, agreement here is much lower at about 30\%, making large divergences in $\Delta$win-rate more common across features.

A selection of features for \texttt{legaladvice} is displayed in Figure~\ref{fig:legaladvice}, with full visualizations for both domains in Appendix~\ref{app:feature-analysis}. One of the most consistent differences observed between Reddit users and \texttt{gpt-5.1} is the tendency for humans to highly rate bot responses pointing to external resources, while \texttt{gpt-5.1} strongly disfavors them. We also observe a pattern of humans preferring responses that advocate self-directed action, while \texttt{gpt-5.1} is more cautious: for the two features \textit{provides detailed, context-specific advice and analysis beyond simply telling the user to contact the police} and \textit{focuses on clarifying facts or formal legal/medical processes rather than suggesting self-help measures like surveillance technology or lawsuits}, for example, the $\Delta$win-rate for \texttt{gpt-5.1} is about 16 percentage points higher than for Reddit users. On the \texttt{askacademia} dataset, meanwhile, we find that many of the generated features proxy wordiness and formality of responses; we find that in general humans tend to prefer more concise and informal comments, while \texttt{gpt-5.1} favors longer and more formal ones.

\section{Conclusion}

In this paper, we showed how automated concept extraction techniques can identify interpretable preference axes for LLM judges. This method validates known biases and uncovers new ones on both general and domain-specific datasets. Directions for future work include optimizing along the Pareto frontier between Interpretability and Predictiveness, further analysis of preference patterns on varied datasets, and normative analysis of when models with certain preference patterns should be favored as judges on particular tasks.

\bibliography{iclr2026_conference}
\bibliographystyle{iclr2026_conference}

\clearpage
\FloatBarrier
\appendix
\section{Appendix}

\subsection{Model Agreement and Position Bias}\label{app:model-agreement}

\begin{table}[t]

\caption{Model agreement and bias statistics. Models: GPT = \texttt{gpt-5.1}, Claude = \texttt{claude-sonnet-4.5}, Gemini = \texttt{gemini-3-flash-preview}.}
\label{tab:model-agreement}

\begin{center}
\begin{tabular}{lccc}
\multicolumn{1}{c}{\bf Metric} &
\multicolumn{1}{c}{\bf GPT} &
\multicolumn{1}{c}{\bf Claude} &
\multicolumn{1}{c}{\bf Gemini}
\\ \hline \\
Agreement with human (\%)                & 70.0 & 68.0 & 70.5 \\
Avg.\ agreement with other LLMs (\%)     & 81.9 & 81.8 & 82.6 \\
Response B preference rate (\%)          & 53.1 & 61.2 & 52.1 \\
Position bias $p$-value                  & $3.87\times10^{-20}$ & $2.76\times10^{-250}$ & $5.82\times10^{-10}$ \\
\end{tabular}
\end{center}

\end{table}

All three models used agree with humans and with each other at similar rates. The rate of agreement with humans is around 70\%. While the position of responses is randomized, such that humans prefer Response B (the second response shown) almost exactly 50\% of the time, all three models demonstrate position bias toward Response B at statistically significant rates. These findings echo previous results \citep{zheng2023judgingllmasajudgemtbenchchatbot}. \texttt{claude-sonnet-4.5} exhibits particularly egregious position bias, preferring Response B over 60\% of the time.

\subsection{Method Details}\label{app:method-details}

\paragraph{Data preparation.} Several preprocessing steps are applied: we randomly deduplicate rows with identical prompts; remove non-English conversations; retain only the first turn of multiturn conversations; and, following \citet{huang2025valueswilddiscoveringanalyzing}, remove prompts that require an objectively correct answer. This final step is necessary because, in cases where an objective answer is required, the primary preference criterion will be whether or not the answer is correct, which is unlikely to be analyzable by concept extraction techniques. The resulting dataset has 27,734 entries.

The prompt used for generating pairwise judgments is borrowed from \citet{judgebench2024} and appears in Appendix~\ref{app:prompts}. Upon generating these judgments, we are immediately able to validate some previous results, as shown in Table~\ref{tab:model-agreement}. In particular, all three LLMs agree with humans at a lower rate than they do with each other, typically around 70\% versus 80\%, and all three exhibit statistically significant position bias, preferring Response B (the second response shown) at higher rates despite the responses being randomized; \texttt{claude-sonnet-4.5} is the most egregious in this second respect, preferring Response B over 60\% of the time. These findings echo well-known results in the LLM-as-a-judge literature, e.g. \citet{zheng2023judgingllmasajudgemtbenchchatbot}.

\subsubsection{Supervised model details (concept bottleneck baselines)}
\label{app:supervised-model-details}

\paragraph{Input representation.}
Let $p\in\mathbb{R}^{d}$ be the prompt embedding and $r\in\mathbb{R}^{d}$ be a response embedding (we use $d=1536$ from \texttt{text-embedding-3-small}). For each (prompt, response) pair, we build the classifier input
$\phi(p,r)=\left[p;\, r;\, p\odot r;\, |p-r|\right]\in\mathbb{R}^{4d}$ by concatenating the prompt embedding, response embedding, elementwise product, and elementwise absolute difference.

\paragraph{Architecture.}
We train a small MLP to score each response independently. The network applies LayerNorm to the $4d$-dimensional input, followed by two hidden layers with GELU activations and dropout, and a linear penultimate layer that serves as the concept bottleneck. Concretely, we use hidden sizes 512 and 128, dropout 0.5, and a 32-dimensional penultimate layer (the ``concept layer''). A final linear head maps this penultimate representation to a single scalar score.

\paragraph{Pairwise training objective.}
Given a prompt with responses $r_0$ and $r_1$, the model produces scalar scores $s_0=f(\phi(p,r_0))$ and $s_1=f(\phi(p,r_1))$ and defines the preference logit as $\ell=s_1-s_0$. We optimize binary cross-entropy with logits (BCEWithLogitsLoss) against the target label $y\in\{0,1\}$ indicating whether response 1 is preferred.

\paragraph{Optimization and evaluation.}
We shuffle and split the dataset into 80\% train and 20\% test. We train with AdamW (learning rate $10^{-3}$, weight decay $5\times 10^{-2}$), batch size 256, for up to 30 epochs with early stopping (patience 5 based on held-out loss). For context, we also report a simple logistic regression baseline on the same engineered embedding features.

\subsubsection{Predictiveness metric}
\paragraph{Definition.}
For each method, we measure \emph{Predictiveness} as the ROC-AUC of a logistic regression classifier trained to predict the LLM judge preference label from that method's full feature vector (i.e. using all learned features, not only those deemed significant).

\paragraph{Procedure.}
We take the matrix of feature values $X\in\mathbb{R}^{n\times d}$ and the corresponding LLM preference labels $y$, dropping examples with missing/invalid labels (e.g. $y=-1$). We then use an 80/20 train/test split, fit an $\ell_2$-regularized scikit-learn \texttt{LogisticRegression} model (\texttt{max\_iter}=1000) on the training set, and report ROC-AUC on the test set using the predicted probabilities.

\subsubsection{Length-controlled $\Delta$win-rate metric}
\label{app:delta-winrate}

\paragraph{Motivation.}
Individual feature activations are often correlated with superficial properties of responses, most notably length. To report an interpretable effect size that isolates the directional association between a feature and preference while controlling for length, we define a length-controlled \emph{delta win-rate} ($\Delta$win-rate) for each feature.

\paragraph{Setup.}
Let $y_i\in\{0,1\}$ denote whether response~1 is preferred over response~0 for example $i$, let $z_{ij}\in\mathbb{R}$ be the (unstandardized) activation of feature $j$, and let
$x_i$ be a standardized control variable given by the difference in word count between responses ($\text{len}(r_1)-\text{len}(r_0)$), normalized to mean~0 and variance~1.

\paragraph{Sign-split logistic model.}
For feature~$j$, we restrict to examples with nonzero activation ($z_{ij}\neq 0$) and define a binary indicator
\[
D_{ij} = \mathbb{I}[z_{ij} > 0],
\]
which captures the sign of the feature activation. We then fit a logistic regression
\[
\Pr(y_i = 1 \mid D_{ij}, x_i)
= \sigma\!\left(\alpha_j + \beta_j D_{ij} + \gamma_j x_i\right),
\]
where $\sigma(\cdot)$ is the logistic sigmoid. This model estimates the effect of a positive (vs.\ negative) activation of feature~$j$ while linearly controlling for length differences.

\paragraph{Delta win-rate.}
We define the $\Delta$win-rate for feature~$j$ as the average change in predicted win probability when flipping the feature sign from negative to positive, holding the observed length control fixed:
\[
\Delta_j
= \frac{1}{|\mathcal{I}_j|}
\sum_{i\in\mathcal{I}_j}
\left[
\sigma(\alpha_j + \beta_j + \gamma_j x_i)
-
\sigma(\alpha_j + \gamma_j x_i)
\right],
\]
where $\mathcal{I}_j=\{i : z_{ij}\neq 0\}$. Intuitively, $\Delta_j$ measures how much more likely response~1 is to be preferred when feature~$j$ is positive rather than negative, after accounting for response length.

\paragraph{Uncertainty estimation.}
We estimate confidence intervals for $\Delta_j$ using nonparametric bootstrap resampling over examples (1{,}000 replicates). For each bootstrap sample, we recompute the standardized length control, refit the sign-split logistic model, and recompute $\Delta_j$. We report the 5th and 95th percentiles of the bootstrap distribution as a 90\% confidence interval.

\subsection{Prompts}\label{app:prompts}

\subsubsection{Feature description prompt}
We use the same feature-description prompt as in Figure~9 of \citet{movva2025whatshumanfeedbacklearning}.

\subsubsection{Fidelity evaluation prompt}
For fidelity evaluation, we prompt the evaluator model with the following instruction template, adapted from \citet{zhong2025explainingdatasetswordsstatistical}:
\begin{lstlisting}
Check which of the two TEXTS exhibits a PROPERTY more strongly. Respond with "A", "B", or "Neither". Do not output any explanation or extra text. Respond with either "A" or "B" when possible, but if you are truly unable to decide, respond with "Neither".

Example 1:
PROPERTY: "mentions a natural scene."
TEXT A: "I love the way the sun sets in the evening."
TEXT B: "I like chairs."
Output: A

Example 2:
PROPERTY: "writes in a 1st person perspective."
TEXT A: "Jacob is smart."
TEXT B: "I think I am smart."
Output: B

Example 3:
PROPERTY: "mentions that the breakfast is good on the airline."
TEXT A: "The airline staff was nice."
TEXT B: "The breakfast on the airline was great."
Output: B

Example 4:
PROPERTY: "uses formal academic language."
TEXT A: "This study investigates nominalization."
TEXT B: "This looks cool."
Output: A

Now complete the following example: respond only with A, B, or Neither.

PROPERTY: {{ interpretation }}
TEXT A: {{ response_0 }}
TEXT B: {{ response_1 }}
Output:
\end{lstlisting}

\subsubsection{LLM-as-a-judge pairwise evaluation prompt}
We use the vanilla JudgeBench pairwise preference prompt \citep{judgebench2024}.

\subsubsection{Subjective prompt filter}
To filter prompts that require an objectively correct answer, we use the following binary classifier prompt adapted from \citet{huang2025valueswilddiscoveringanalyzing}:
\begin{lstlisting}
The following is a single-turn exchange between a user and an AI assistant:

{{transcript}}

<question>
Does the assistant's reply require giving a subjective judgment?

Answer "Yes" if the request instead invites opinion, preference, creative generation, advice, or other context-dependent interpretation.
Answer "No" if the user's request calls for a verifiable or factually correct response (e.g., definitions, calculations, technical explanations, factual information).

If the distinction is unclear, decide based on what type of answer would best satisfy the user's prompt.
Output only <answer>Yes</answer> or <answer>No</answer>.
</question>
\end{lstlisting}

\subsection{Feature Analysis}\label{app:feature-analysis}

\begin{figure}[H]
\begin{center}
\includegraphics[width=\linewidth]{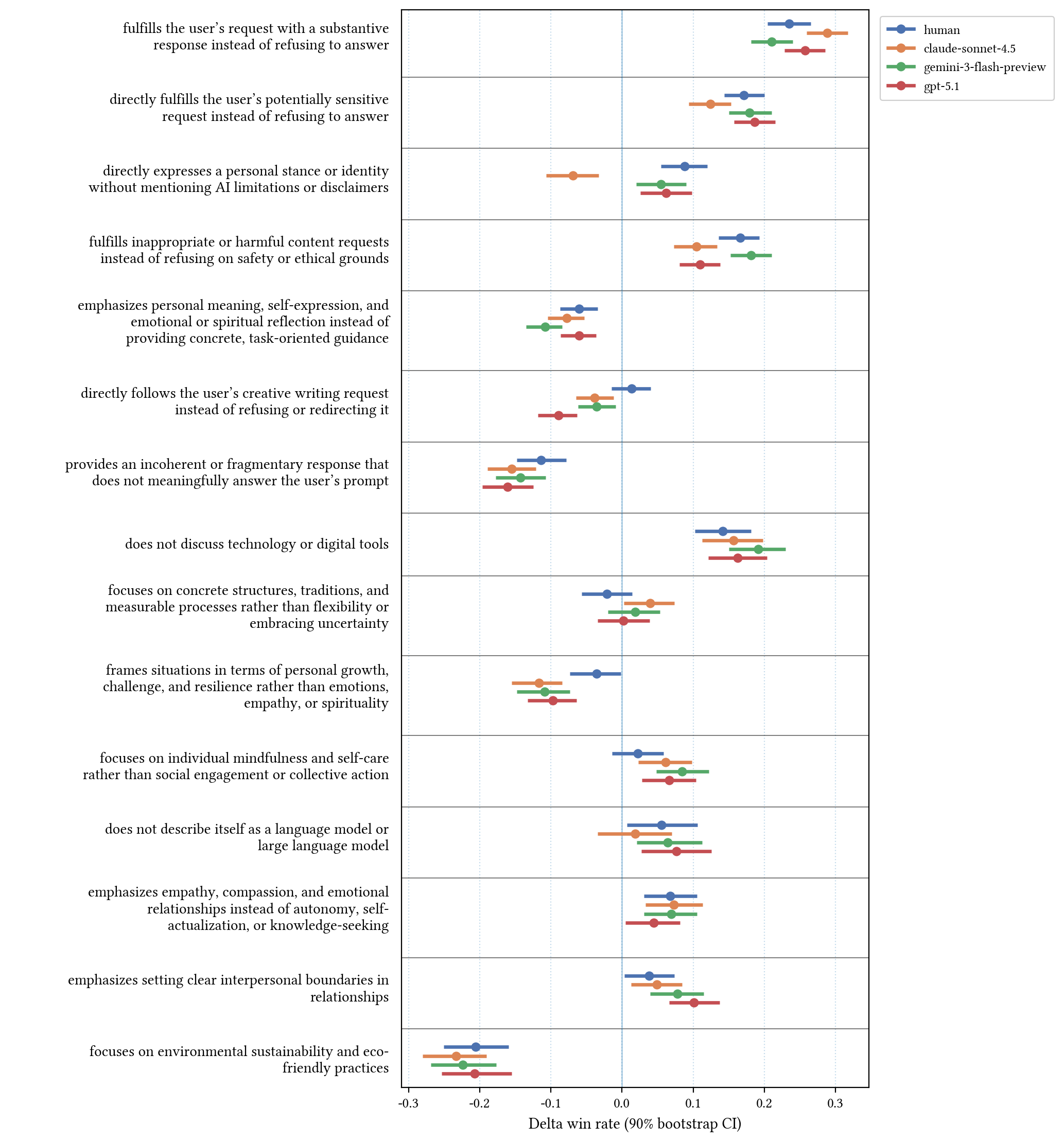}
\end{center}
\caption{All Differential SAE features for the combined preference dataset, with interpretations and $\Delta$win-rate for human and LLM annotators.}
\label{fig:combined-full}
\end{figure}

Figure~\ref{fig:combined-full} shows the full set of Differential SAE features on the combined preference dataset.

\begin{figure}[t]
\begin{center}
\includegraphics[width=\linewidth]{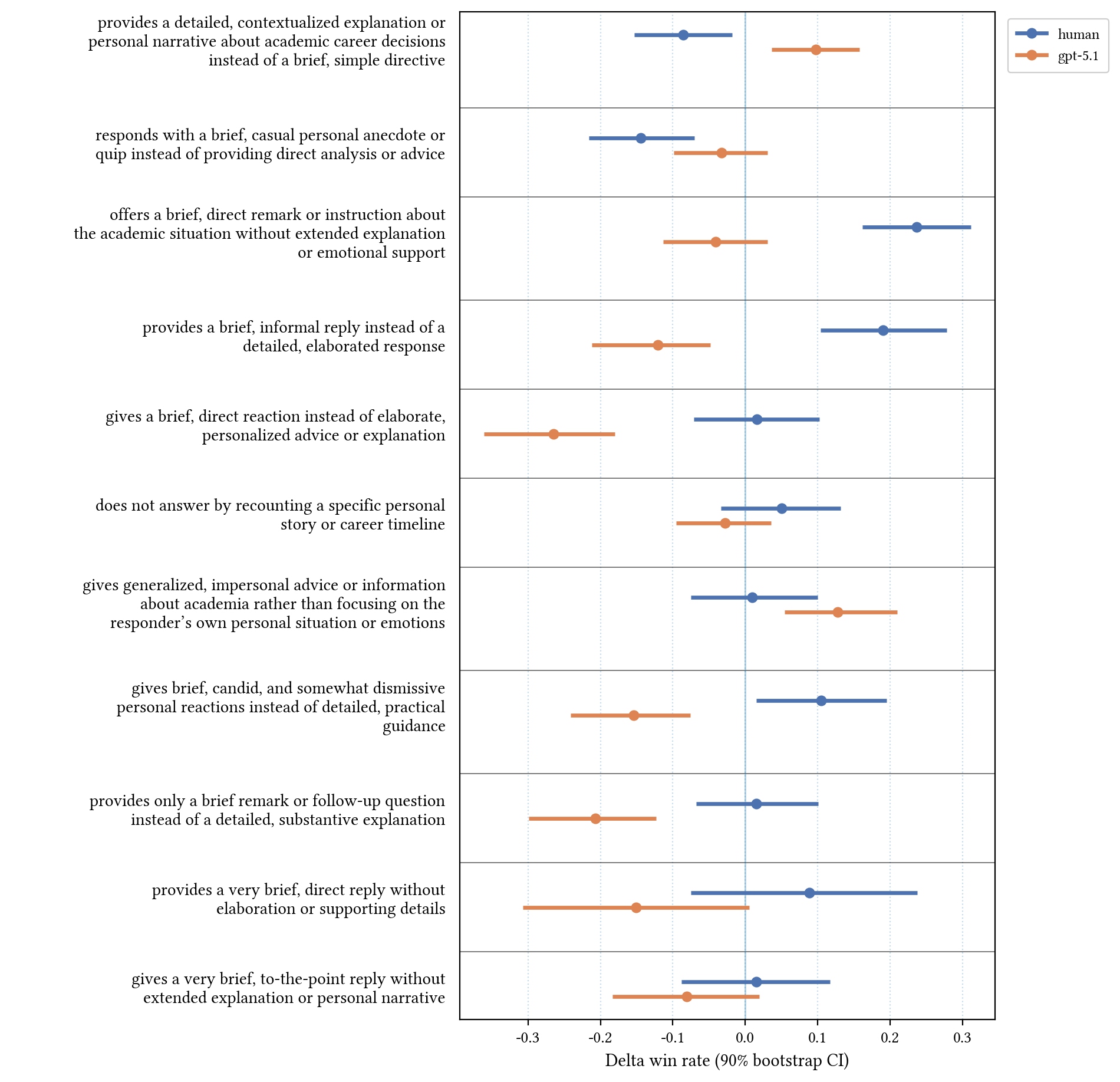}
\end{center}
\caption{All features for the \texttt{askacademia} dataset.}
\label{fig:askacademia-full}
\end{figure}

Figure~\ref{fig:askacademia-full} shows the full feature visualization for \texttt{askacademia}.

\begin{figure}[t]
\begin{center}
\includegraphics[width=\linewidth]{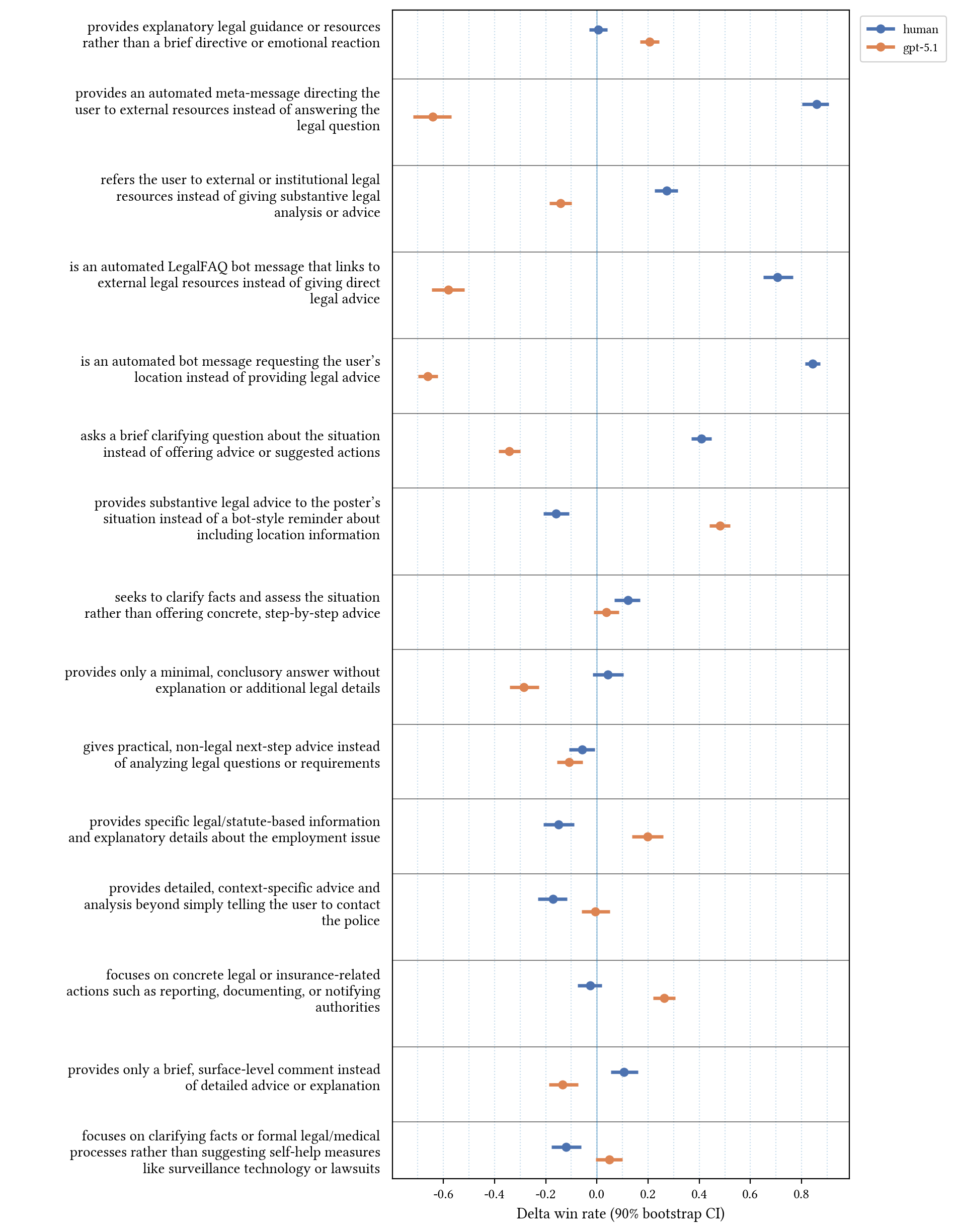}
\end{center}
\caption{All features for the \texttt{legaladvice} dataset.}
\label{fig:legaladvice-full}
\end{figure}

Figure~\ref{fig:legaladvice-full} shows the full feature visualization for \texttt{legaladvice}.

\end{document}